\documentclass[runningheads]{llncs}
\usepackage{graphicx}
\usepackage{amsmath,amssymb} 
\usepackage{graphicx}
\usepackage{color}\usepackage[width=122mm,left=12mm,paperwidth=146mm,height=193mm,top=12mm,paperheight=217mm]{geometry}
\begin{document}

\newcommand{\fig}[1]{Figure~\ref{fig:#1}}
\newcommand{\sect}[1]{Section~\ref{sect:#1}}
\newcommand{\tab}[1]{Table~\ref{tab:#1}}
\newcommand{\eq}[1]{(\ref{eq:#1})}

\newcommand{\Lfi}[1]{\ensuremath{L^5(#1)}}
\newcommand{\Lsi}[1]{\ensuremath{L^6(#1)}}
\newcommand{\Lse}[1]{\ensuremath{L^7(#1)}}
\newcommand{\Lei}[1]{\ensuremath{L^8(#1)}}

\renewcommand{\topfraction}{0.85}
\renewcommand{\textfraction}{0.1}
\renewcommand{\floatpagefraction}{0.85}

\pagestyle{headings}
\mainmatter
\title{Neural Codes for Image Retrieval} 

\titlerunning{Neural Codes for Image Retrieval}

\authorrunning{A. Babenko, A. Slesarev, A. Chigorin, V. Lempitsky}

\author{Artem Babenko$^{1,3}$, 
        Anton Slesarev$^1$, 
        Alexandr Chigorin$^1$, 
        Victor Lempitsky$^2$}

\institute{$^1$Yandex, \qquad $^2$Skolkovo Institute of Science and Technology (Skoltech), \\$^3$Moscow Institute of Physics and Technology}

\maketitle

\begin{abstract}
It has been shown that the activations invoked by an image within the top layers of a large convolutional neural network provide a high-level descriptor of the visual content of the image. In this paper, we investigate the use of such descriptors (neural codes) within the image retrieval application. In the experiments with several standard retrieval benchmarks, we establish that neural codes perform competitively even when the convolutional neural network has been trained for an unrelated classification task (e.g.\ Image-Net). We also evaluate the improvement in the retrieval performance of neural codes, when the network is retrained on a dataset of images that are similar to images encountered at test time.

We further evaluate the performance of the compressed neural codes and show that a simple PCA compression provides very good short codes that give state-of-the-art accuracy on a number of datasets. In general, neural codes turn out to be much more resilient to such compression in comparison other state-of-the-art descriptors. Finally, we show that discriminative dimensionality reduction trained on a dataset of pairs of matched photographs  improves the performance of PCA-compressed neural codes even further. Overall, our quantitative experiments demonstrate the promise of neural codes as visual descriptors for image retrieval.


\end{abstract}

\section{Introduction}

Deep convolutional neural networks~\cite{LeCun89} have recently advanced the state-of-the-art in image classification dramatically~\cite{Krizhevsky12} and have consequently attracted a lot of interest within the computer vision community. A separate but related to the image classification problem is the problem of image retrieval, i.e.\ the task of finding images containing the same object or scene as in a query image. It has been suggested that the features emerging in the upper layers of the CNN learned to classify images can serve as good descriptors for image retrieval. In particular, Krizhevsky et al.~\cite{Krizhevsky12} have shown some qualitative evidence for that. Here we interesed in establishing the quantitative performance of such features (which we refer to as \emph{neural codes}) and their variations.

We start by providing a quantitative evaluation of the image retrieval performance of the features that emerge within the convolutional neural network trained to recognize Image-Net~\cite{ILSVRC} classes. We measure such performance on four standard benchmark datasets: INRIA Holidays~\cite{Holidays}, Oxford Buildings, Oxford Building 105K~\cite{Philbin07}, and the University of Kentucky benchmark (UKB)~\cite{Nister06}. Perhaps unsurprisingly, these deep features perform well, although not better than other state-of-the-art holistic features (e.g.\ Fisher vectors). Interestingly, the relative performance of different layers of the CNN varies in different retrieval setups, and the best performance on the standard retrieval datasets is achieved by the features in the middle of the fully-connected layers hierarchy.

\begin{figure*}
\centering
\includegraphics[width=13cm]{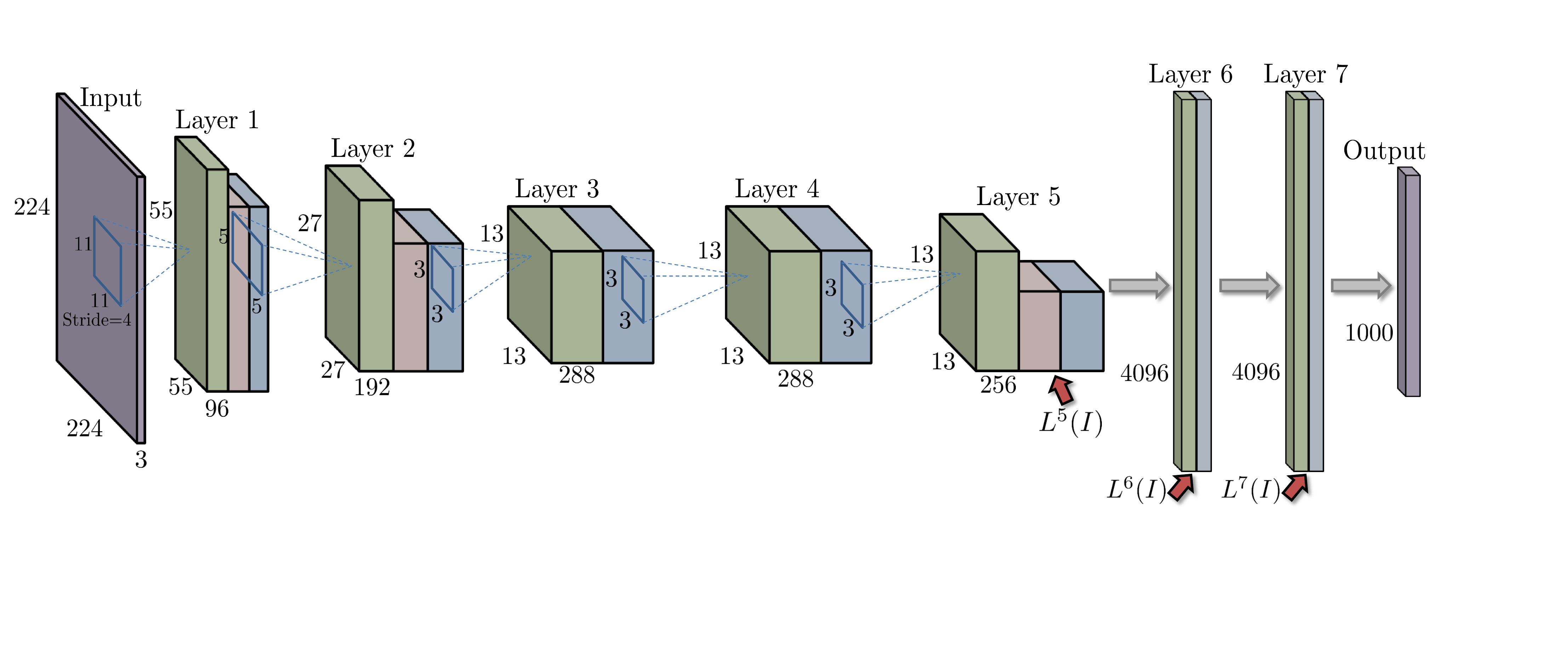}
\vspace{-2cm}
\caption{The convolutional neural network architecture used on our experiments. Purple nodes correspond to input (an RGB image of size $224\times 224$) and output ($1000$ class labels). Green units correspond to outputs of convolutions, red units correspond to the outputs of max pooling, and blue units correspond to the outputs of rectified linear (ReLU) transform. Layers 6, 7, and 8 (the output) are fully connected to the preceding layers. The units that correspond to the neural codes used in our experiments are shown with red arrows. Stride=4 are used in the first convolutional layer, and stride=1 in the rest.}
\label{fig:net_structure}
\end{figure*}

The good performance of neural codes demonstrate their universality, since the task the network was trained for (i.e.\ classifying Image-Net classes) is quite different from  the retrieval task we consider. Despite the evidence of such universality, there is an obvious possibility to improve the performance of deep features by adapting them to the task, and such adaptation is the subject of the second part of the paper. Towards this end, we assemble a large-scale image dataset, where the classes correspond to landmarks (similar to \cite{Li09}), and retrain the CNN on this collection using the original image-net network parameters as initialization. After such training, we observe a considerable improvement of the retrieval performance on the datasets with similar image statistics, such as INRIA Holidays and Oxford Buildings, while the performance on the unrelated UKB dataset degrades. In the second experiment of this kind, we retrain the initial network on the Multi-view RGB-D dataset~\cite{RGBD} of turntable views of different objects. As expected, we observe the improvement on the more related UKB dataset, while the performance on other datasets degrades or stays the same.

Finally, we focus our evaluation on the performance of the compact versions of the neural codes. We evaluate the performance of the PCA compression and observe that neural codes can be compressed very substantially, e.g.\ to 128 dimensions, with virtually no loss of the retrieval accuracy. Overall, the degradation from the PCA compression incurred by the neural codes is considerably smaller than the degradation incurred by other holistic descriptors. This makes the use of neural codes particularly attractive for large-scale retrieval applications, where the memory footprint of a descriptor often represents the major bottleneck.

Pushing the compression to the extreme, to e.g.\ 16 dimensions leads to considerable degradation, as long as PCA is used for the compression. We experiment with discriminative dimensionality reduction learned on an automatically collected large collection of pairs of photos depicting the same object (around 900K pairs). When trained on such a dataset, the discriminative dimensionality reduction performs substantially better than PCA and achieves high retrieval accuracy for very short codes (e.g.\ 0.368 mAP on Oxford Buildings for 16-dimensional features). 

\section{Related work}

Our paper was inspired by the strong performance of convolutional neural networks (CNN) in image classification tasks, and the qualitative evidence of their feasibility for image retrieval provided in~\cite{Krizhevsky12}. A subsequent report~\cite{Donahue13} demonstrated that features emerging within the top layers of large deep CNNs can be reused for classification tasks dissimilar from the original classification task. Convolutional networks have also been used to produce descriptors suitable for retrieval within the \emph{siamese architectures} \cite{Chopra05}.

In the domain of ``shallow'' architectures, there is a line of works on applying the responses of discriminatively trained multiclass classifiers as descriptors within retrieval applications. Thus, \cite{Wang09} uses the output of classifiers trained to predict membership of Flickr groups as image descriptors. Likewise, very compact descriptors based on the output of binary classifiers trained for a large number of classes (\emph{classemes}) were proposed in \cite{Torresani10}. Several works such as \cite{Kumar09} used the outputs of discriminatively trained classifiers to describe human faces, obtaining high-performing face descriptors.

The current state-of-the-art holistic image descriptors are obtained by the aggregation of local gradient-based descriptors. Fisher Vectors~\cite{Perronnin10} is the best known descriptor of this kind, however its performance has been recently superceded by the triangulation embedding suggested in \cite{Jegou14} (another recent paper~\cite{Tolias13} have introduced descriptors that can also achieve very high performance, however the memory footprint of such descriptors is at least an order of magnitude larger than uncompressed Fisher vectors, which makes such descriptors unsuitable for most applications). 

In~\cite{Gordo12}, the dimensionality reduction of Fisher vectors is considered, and it is suggested to use Image-Net  to discover discriminative low-dimensional subspace. The best performing variant of such dimensionality reduction in~\cite{Gordo12} is based on adding a hidden unit layer and a classifier output layer on top of Fisher vectors. After training on a subset of Image-Net, the low-dimensional activations of the hidden layer are used as descriptors for image retrieval. The architecture of \cite{Gordo12} therefore is in many respects similar to those we investigate here, as it is deep (although not as multi-layered as in our case), and is trained on image-net classes. Still, the representations derived in \cite{Gordo12} are based on hand-crafted features (SIFT and local color histograms) as opposed to neural codes derived from CNNs that are learned from the bottom up.

There is also a large body of work on dimensionality reduction and metric learning \cite{Yang2006}. In the last part of the paper we used a variant of the discriminative dimensionality reduction similar to \cite{Simonyan13}.

Independently and in parallel with our work, the use of neural codes for image retrieval (among other applications) has been investigated in \cite{Razavian14}. Their findings are largely consistent with ours, however there is a substantial difference from this work in the way the neural codes are extracted from images. Specifically, \cite{Razavian14} extract a large number of neural codes from each image by applying a CNN in a ``jumping window'' manner. In contrast to that, we focus on holistic descriptors where the whole image is mapped to a single vector, thus resulting in a substantially more compact and faster-to-compute descriptors, and we also investigate the performance of compressed holistic descriptors. 

Furthermore, we investigate in details how retraining of a CNN on different datasets impact the retrieval performance of the corresponding neural codes. Another concurrent work~\cite{Oquab14} investigated how similar retraining can be used to adapt the Image-Net derived networks to smaller classification datasets.

\section{Using Pretrained Neural Codes}

{\bf Deep convolutional architecture.} In this section, we evaluate the performance of neural codes obtained by passing images through a deep convolution network, trained to classify 1000 Image-Net classes \cite{Krizhevsky12}. In particular, we use our own reimplementation of the system from \cite{Krizhevsky12}. The model includes five convolutional layers, each including a convolution, a rectified linear (ReLU) transform ($f(x) = \max(x,0)$), and a max pooling transform (layers 1, 2, and 2). At the top of the architecture are three fully connected layers (``layer 6'', ``layer 7'', ``layer 8''), which take as an input the output of the previous layer, multiply it by a matrix, and, in the case of layers 6, and 7 applies a rectified linear transform. The network is trained so that the layer 8 output corresponds to the one-hot encoding of the class label. The softmax loss is used during training. The results of the training on the ILSVRC dataset \cite{ILSVRC} closely matches the result of a single CNN reported in \cite{Krizhevsky12} (more precisely, the resulting accuracy is worse by 2\%).
Our network architecture is schematically illustrated on \fig{net_structure}.

The network is applicable to $224\times 224$ images. Images of other dimensions are resized to $224\times 224$ (without cropping). The CNN architecture is feed-forward, and given an image $I$, it produces a sequence of layer activations. We denote with \Lfi{I}, \Lsi{I}, and \Lse{I} the activations (output) of the corresponding layer \emph{prior} to the ReLU transform. Naturally, each of these high-dimensional vectors represent a \emph{deep} descriptor (a \emph{neural code}) of the input image. 

{\bf Benchmark datasets.} We evaluate the performance of neural codes on four standard datasets listed below. The results for top performing methods based on holistic descriptors (of dimensionality upto $32$K) are given in \tab{uncompressed}.

{\em Oxford Buildings dataset~\cite{Philbin07} (Oxford).} The dataset consists of 5062 photographs collected from Flickr and corresponding to major Oxford landmarks. Images corresponding to 11 landmarks (some having complex structure and comprising several buildings) are manually annotated. The 55 hold-out queries evenly distributed over those 11 landmarks are provided, and the performance of a retrieval method is reported as a mean average precision (mAP) \cite{Philbin07} over the provided queries.

{\em Oxford Buildings dataset+100K~\cite{Philbin07} (Oxford 105K).} The same dataset with the same associated protocol, but with additional 100K distractor images provided by the dataset authors.

{\em INRIA Holidays dataset~\cite{Holidays} (Holidays).} The dataset consists of 1491 vacation photographs corresponding to 500 groups based on same scene or object. One image from each group serves as a query. The performance is reported as mean average precision over 500 queries. Some images in the dataset are not in a natural orientation (rotated by $\pm 90$ degrees). As deep architectures that we consider are trained on the images in a normal orientation, we follow several previous works, and manually bring all images in the dataset to the normal orientation. In a sequel, all our results are for this modified dataset. We also experimented with an unrotated version and found the performance in most settings to be worse by about $0.03$ mAP. Most of the performance drop can be regained back using data augmentation (rotating by $\pm90$) on the dataset and on the query sides. 

{\em University of Kentucky Benchmark dataset ~\cite{Nister06} (UKB).} The dataset includes 10,200 indoor photographs of 2550 objects (4 photos per object). Each image is used to query the rest of the dataset. The performance is reported as the average number of same-object images within the top-4 results, and is a number between 0 and 4.

{\bf Results.} The results for neural codes produced with a network trained on ILSVRC classes are given in the middle part of \tab{uncompressed}. All results were obtained using L2-distance on L2-normalized neural codes. We give the results corresponding to each of the layers 5, 6, 7. We have also tried the output of layer 8 (corresponding to the ILSVRC class probabilities and thus closely related to previous works that used class probabilities as descriptors), however it performed considerably worse (e.g.\ 0.02 mAP worse than layer 5 on Holidays).

\begin{figure*}
\centering
\begin{tabular}{cc}
\Lfi{I}&\includegraphics[width=11.5cm]{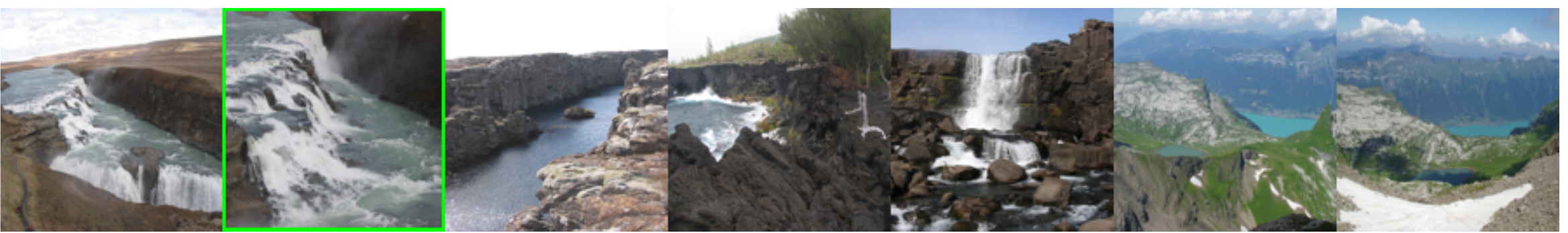}\\
\Lsi{I}&\includegraphics[width=11.5cm]{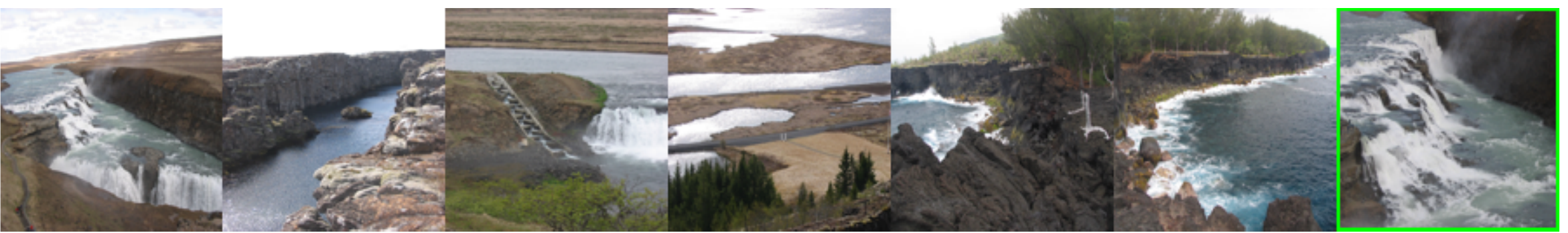}\\
\Lse{I}&\includegraphics[width=11.5cm]{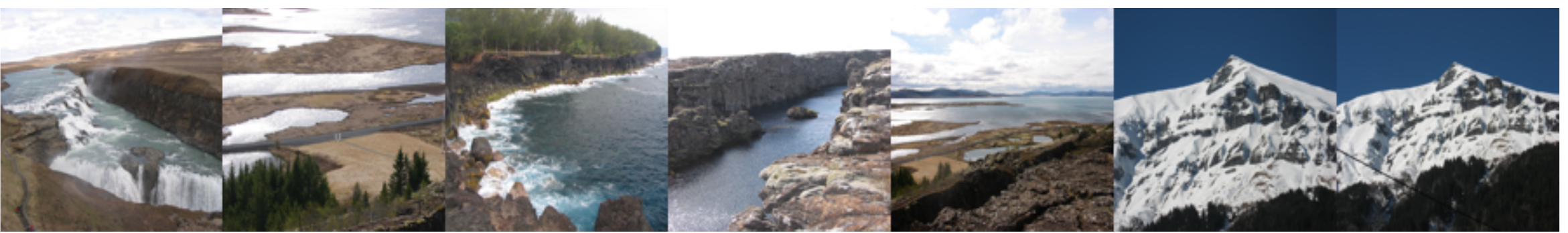}\\
\end{tabular}
\caption{A retrieval example on Holidays dataset where Layer 5 gives the best result among other layers, presumably because of its reliance on relatively low-level texture features rather than high level concepts.  The left-most image in each row corresponds to the query, correct answers are outlined in green.}
\label{fig:good5}
\end{figure*}

\begin{figure*}
\centering
\begin{tabular}{cc}
\Lfi{I}&\includegraphics[width=11.5cm]{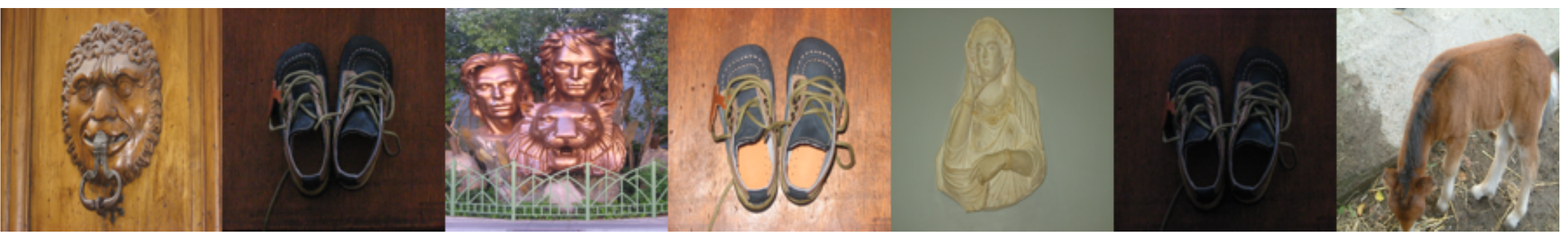}\\
\Lsi{I}&\includegraphics[width=11.5cm]{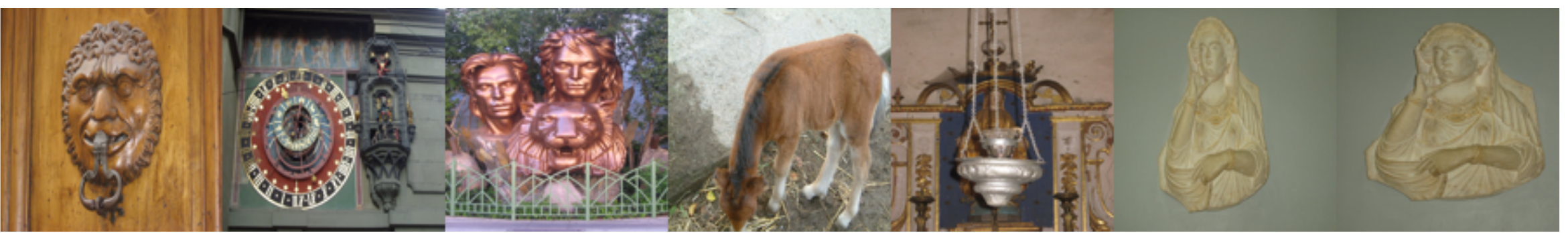}\\
\Lse{I}&\includegraphics[width=11.5cm]{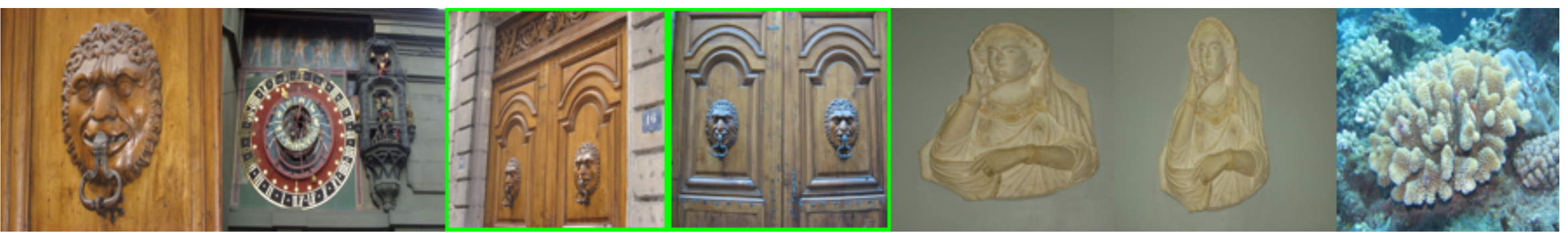}\\
\end{tabular}
\caption{A retrieval example on Holidays dataset where Layer 7 gives the best result among other layers, presumably because of its reliance on high level concepts.  The left-most image in each row corresponds to the query, correct answers are outlined in green.}
\label{fig:good7}
\end{figure*}

Among all the layers, the 6th layer performs the best, however it is not uniformly better for all queries (see \fig{good5} and \fig{good7}). Still, the results obtained using simple combination of the codes (e.g.\ sum or concatenation) were worse than \Lsi{I}-codes alone, and more complex non-linear combination rules we experimented with gave only marginal improvement.

Overall, the results obtained using \Lsi{I}-codes are in the same ballpark, but not superior compared to state-of-the-art. Their strong performance is however remarkable given the disparity between the ILSVRC classification task and the retrieval tasks considered here. 

\section{Retrained neural codes}

A straightforward idea for the improvement of the performance of neural codes is to retrain the convolutional architecture on the dataset with image statistics and classes that are more relevant for datasets considered at test time.

{\bf The Landmarks dataset.} We first focus on collecting the dataset that is relevant to the landmark-type datasets (Holidays and Oxford Buildings). The collection of such dataset is an untrivial task, and we chose a (semi)-automated approach for that. We start by selecting 10,000 most viewed landmark Wikipedia pages (over the last month). For each page, we used the title of the page as a query to Yandex image search engine\footnote{\tt http://images.yandex.ru}, and then downloaded 1000 top images returned in response to the query (or less, if the query returned less images). 

\begin{figure*}
\centering
\begin{tabular}{c}
\includegraphics[width=11cm]{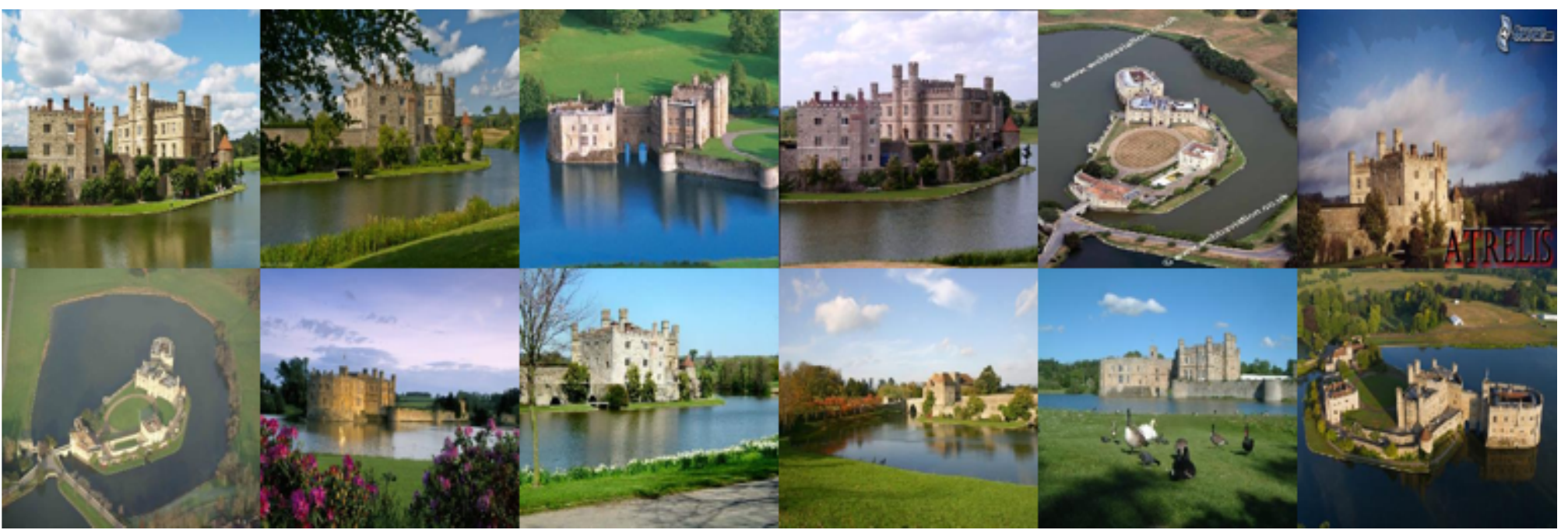}\\
\includegraphics[width=11cm]{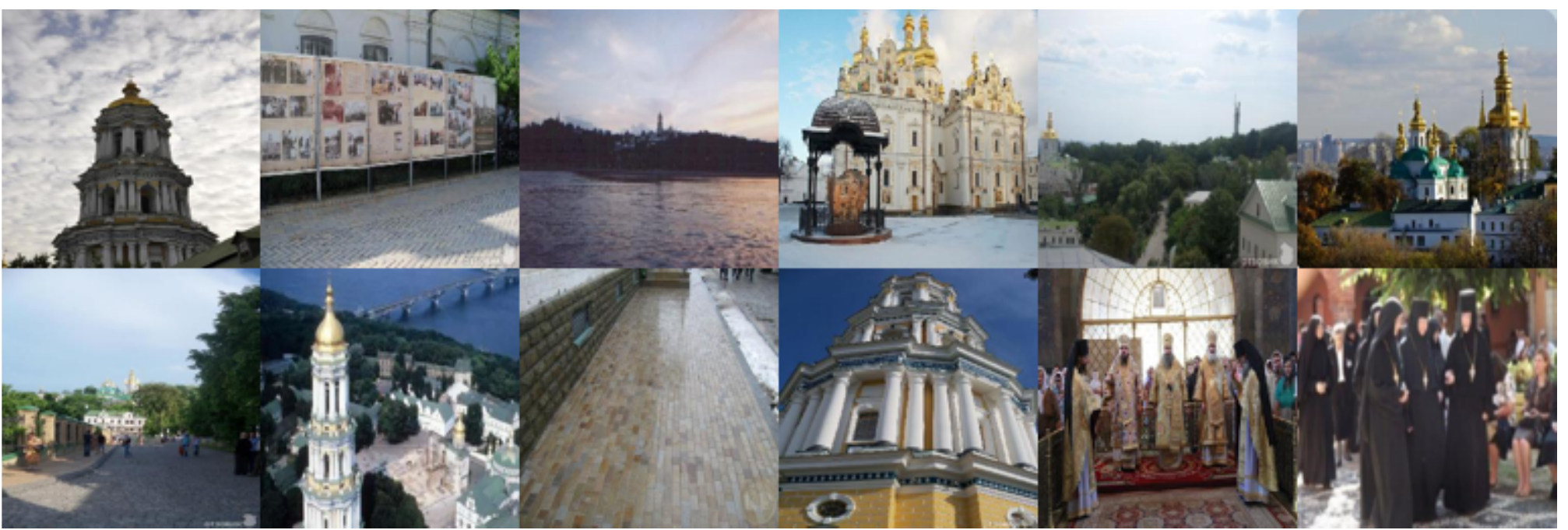}
\end{tabular}
\caption{Sample images from the ''Leeds Castle'' and ''Kiev Pechersk Lavra'' classes of the collected Landmarks dataset. The first class contains mostly ``clean'' outdoor images sharing the same building while the second class contains a lot of  indoor photographs that do not share common geometry with the outdoor photos.}
\label{fig:class_examples}
\end{figure*}

At the second stage, we eyeballed the returned images by looking at the hundred of photographs from the top of the response and at an another hundred sampled uniformly from the remaining images (900 or less). We then manually classify the downloaded list into one of the following three classes: (1) ``take all'' (at least 80\% in both hundreds are relevant, i.e.\ are actual photographs of the landmark), (2) ``take top'' (at least 80\% in the first hundred are relevant, but the second hundred has more than 20\% non-relevant images, including logos, maps, portraits, wrong scenes/objects), (3) ``unsuitable'' (more than 20\% non-relevant images even within the first hundred). Overall, in this way we found 252 ``take all'' classes, and 420 ``take top'' images. \fig{class_examples} shows two typical examples of classes in the collected dataset.
We then assembled the dataset out of these classes, taking either top 1000 images (for ``take all'' classes) or top 100 images (for ``take top'' classes) for each query. Overall the resulting dataset has 672 classes and 213,678 images. During the collection, we excluded queries related to Oxford, and we also removed few near-duplicates with the Holidays dataset from the final dataset. We provide the list of the queries and the URLs at the project webpage\footnote{\tt http://sites.skoltech.ru/compvision/projects/neuralcodes/}.

Our approach for a landmark dataset collection is thus different from that of \cite{Li09} that uses Flickr crawling to assemble a similar dataset in a fully automatic way. The statistics of images indexed by image search engines and of geotagged user photographs is different, so it would be interesting to try the adaptation using the Flickr-crawled dataset.

We then used the collected dataset to train the CNN with the same architecture as for the ILSVRC (except for the number of output nodes that we changed to 672). We initialized our model by the original ILSVRC CNN (again except for the last layer). Otherwise, the training was the same as for the original network.

\begin{table}[t]
\centering
\addtolength{\tabcolsep}{2pt}
\noindent
\begin{tabular}{|c|c|c|c|c|c|}
\hline
Descriptor & Dims & Oxford & Oxford 105K & Holidays & UKB\\
\hline
Fisher+color\cite{Gordo12}      & 4096  &  ---  &  ---  &  \bf{0.774}  & 3.19\\
VLAD+adapt+innorm\cite{arandjelovic13}     & 32768 & 0.555 &  ---  &  0.646  & --- \\
Sparse-coded features\cite{ge2013sparse} & 11024 &  ---  &  ---  &  0.767  & \bf{3.76}\\
Triangulation embedding\cite{Jegou14} & 8064 &  \bf{0.676}  &  \bf{0.611}  &  0.771  & 3.53\\
\hline
\multicolumn{6}{|c|}{\bf Neural codes trained on ILSVRC}\\
\hline
Layer 5 & 9216 & 0.389 &  ---  & 0.690* & 3.09\\
Layer 6 & 4096 & 0.435 & 0.392 & 0.749* & 3.43\\
Layer 7 & 4096 & 0.430 &  ---  & 0.736* & 3.39\\
\hline
\multicolumn{6}{|c|}{\bf After retraining on the Landmarks dataset}\\
\hline
Layer 5 & 9216 & 0.387 &  ---  & 0.674* & 2.99\\
Layer 6 & 4096 & 0.545 & 0.512 & \bf{0.793}* & 3.29\\
Layer 7 & 4096 & 0.538 &  ---  & 0.764* & 3.19\\
\hline
\multicolumn{6}{|c|}{\bf After retraining on turntable views (Multi-view RGB-D)}\\
\hline
Layer 5 & 9216 & 0.348 & --- & 0.682* & 3.13\\
Layer 6 & 4096 & 0.393 & 0.351 & 0.754* & 3.56\\
Layer 7 & 4096 & 0.362 & --- & 0.730* & 3.53\\
\hline
\end{tabular}
\vspace{0.5cm}
\caption{Full-size holistic descriptors: comparison with state-of-the-art (holistic descriptors with the dimensionality up to 32K). The neural codes are competitive with the state-of-the-art and benefit considerably from retraining on related datasets (Landmarks for Oxford Buildings and Holidays; turntable sequences for UKB). $\star$ indicate the results obtained for the rotated version of Holidays, where all images are set into their natural orientation.}
\label{tab:uncompressed}
\end{table}



\begin{figure*}
\centering
\begin{tabular}{c}
\includegraphics[width=12cm]{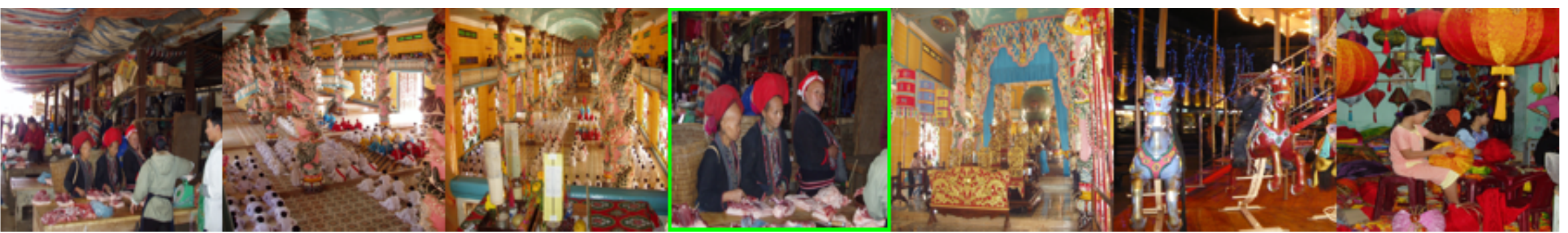}\\
\includegraphics[width=12cm]{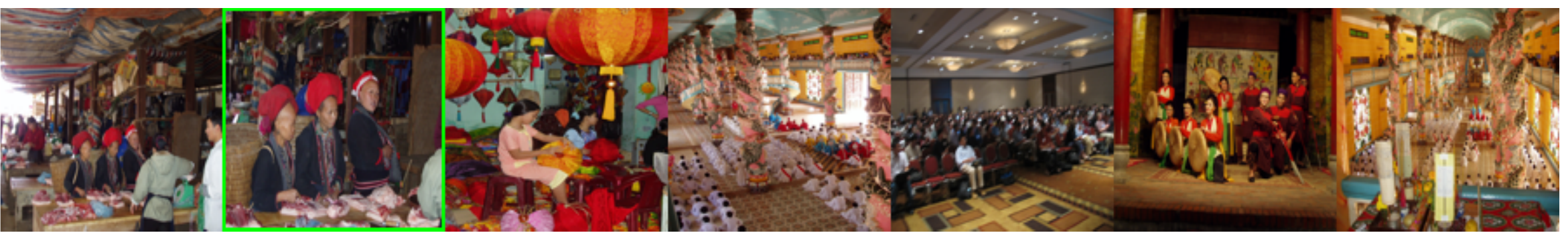}\\
\includegraphics[width=12cm]{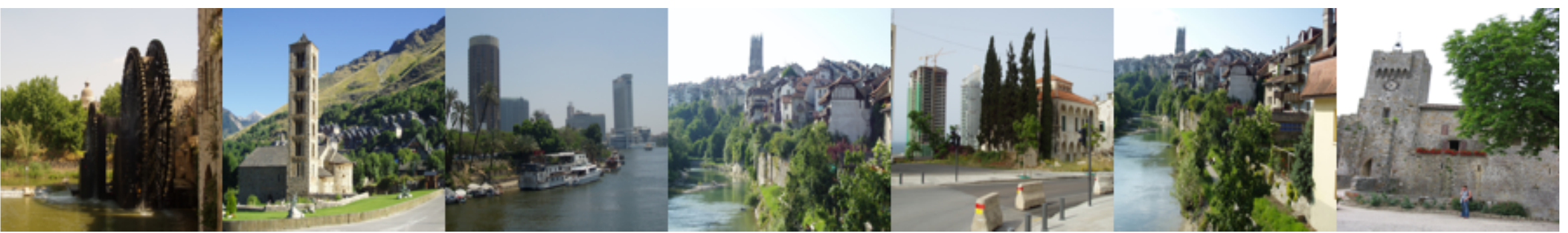}\\
\includegraphics[width=12cm]{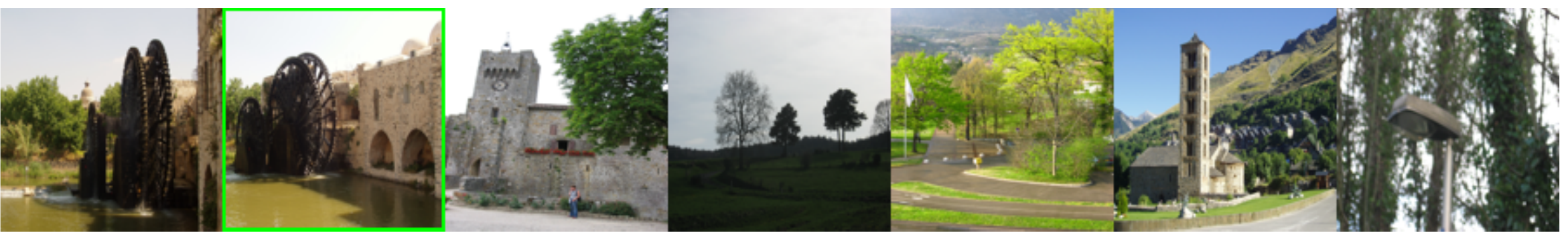}\\
\includegraphics[width=12cm]{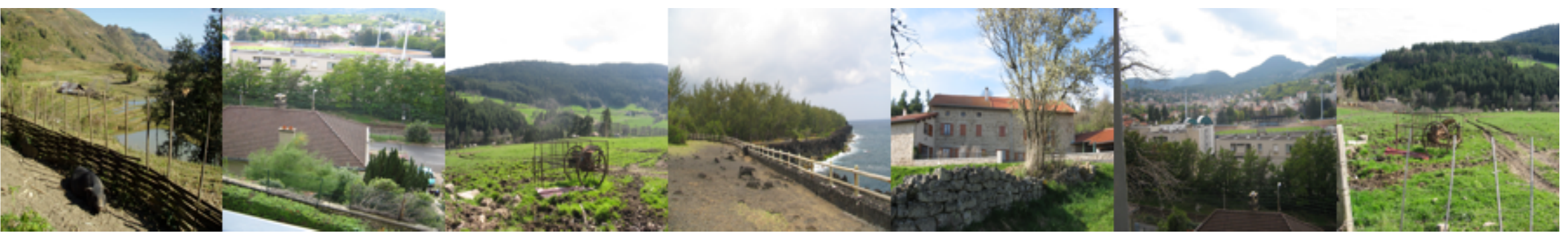}\\
\includegraphics[width=12cm]{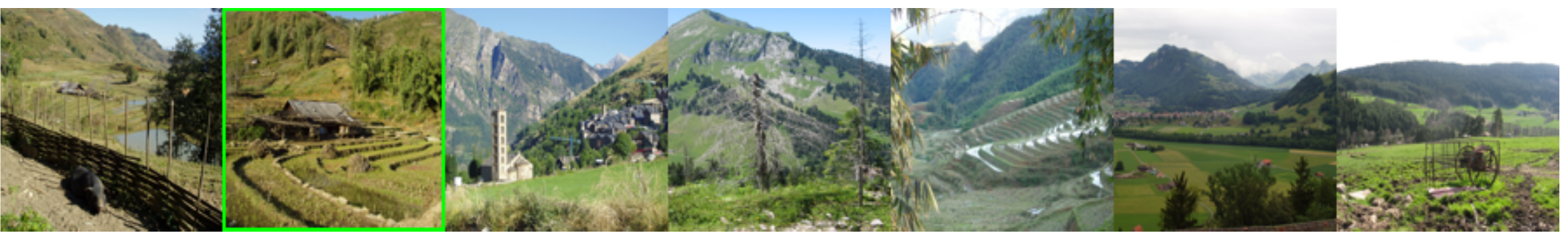}\\
\includegraphics[width=12cm]{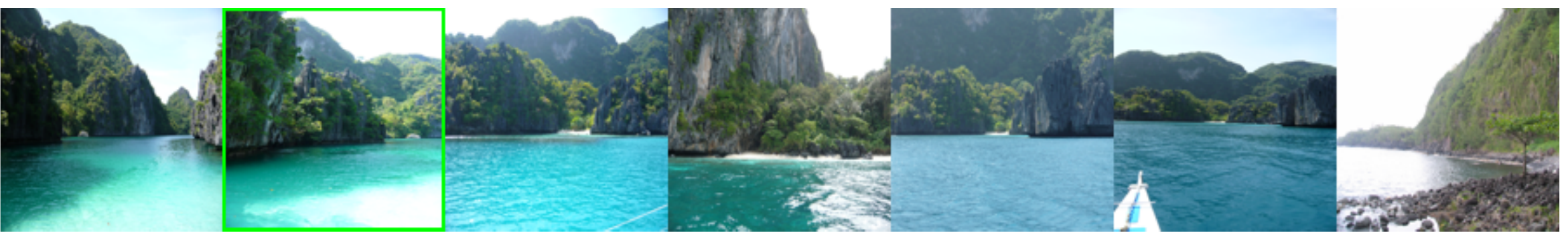}\\
\includegraphics[width=12cm]{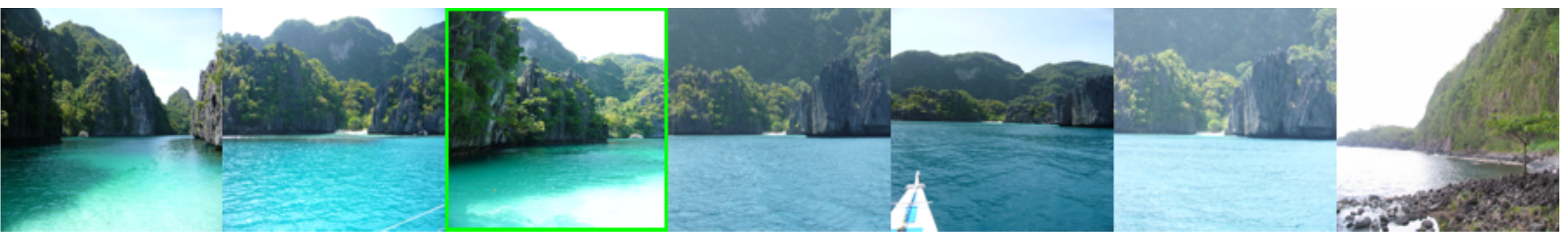}\\
\end{tabular}
\caption{Examples of Holidays queries with large differences between the results of the original and the retrained neural codes (retraining on Landmarks). In each row pair, the left-most images correspond to the query, the top row corresponds to the result with the original neural code, the bottom row corresponds to the retrained neural code. For most  queries, the adaptation by retraining is helpful. The bottom row shows a rare exception.}
\label{fig:adapted_vs_original}
\end{figure*}


{\bf Results for retrained neural codes.} The results for neural codes produced with a network retrained on the landmark classes are given in \tab{uncompressed}. As expected, the difference with respect to the original neural codes is  related to the similarity between the landmark photographs and the particular retrieval dataset. Thus, there is a very big improvement for Oxford and Oxford 105K datasets, which are also based on landmark photographs. The improvement for the Holidays dataset is smaller but still very considerable. The performance of adapted \Lsi{I} features on the Holidays dataset is better then for previously published systems based on holistic features (unless much higher dimensionality as in \cite{Tolias13} is considered). Representative retrieval examples comparing the results obtained with the original and the retrained neural codes are presented in \fig{adapted_vs_original}. We also tried to train a CNN on the landmarks dataset with random initialization (i.e.\ trained from scratch) but observed poor performance due to a smaller number of training images and a higher ratio of irrelevant images compared to ILSVRC.

Interestingly, while we obtain an improvement by retraining the CNN on the Landmarks dataset, no improvement over the original neural codes was obtained by retraining the CNN on the SUN dataset~\cite{SUN10}. Apparently, this is because each SUN class still correspond to {\em different} scenes with the same usage type, while each class in the Landmark dataset as well as in the Holidays and Oxford datasets corresponds to the same object (e.g.\ building).

{\bf Adaptation on the turntable sequences.}  After retraining on the Landmarks collection, the  performance on the UKB dataset drops. This reflects the fact that the classes in the UKB dataset, which correspond to multiple indoor views of different small objects, are more similar to some classes within ILSVRC than to landmark photographs. To confirm this, we performed the second retraining experiment, where we used the Multi-view RGB-D dataset~\cite{RGBD} which contains turntable views of 300 household objects. We treat each object as a separate class and sample 200 images per class. We retrain the network (again, initialized by the ILSVRC CNN) on this dataset of 60,000 images (the depth channel was discarded). Once again, we observed (\tab{uncompressed}) that this retraining provides an increase in the retrieval performance on the related dataset, as the accuracy on the  UKB increased from 3.43 to 3.56. The performance on the unrelated datasets (Oxford, Oxford-105K) dropped.

\section{Compressed neural codes}

As the neural codes in our experiments are high-dimensional (e.g. 4096 for \Lsi{I}), albeit less high-dimensional than other state-of-the-art holistic descriptors, a question of their efficient compression arises. In this section, we evaluate two different strategies for such compression. First, we investigate how efficiency of neural codes degrades with the common PCA-based compression. An important finding is that this degradation is rather graceful. Second, we assess a more sophisticated procedure based on discriminative dimensionality reduction. We focus our evaluation on \Lsi{I}, since the performance of the neural codes associated with the sixth layer was consistently better than with the codes from other layers. 

{\bf PCA compression.} We first evaluate the performance of different versions of neural codes after PCA compression to a different number of dimensions (\tab{compressed}). Here, PCA training was performed on 100,000 random images from the Landmark dataset.

\begin{table}[t]
\centering
\addtolength{\tabcolsep}{4pt}
\noindent
\begin{tabular}{|c|c|c|c|c|c|c|}
\hline
Dimensions & 16 & 32 & 64 & 128 & 256 & 512\\
\hline
\multicolumn{7}{|c|}{\bf Oxford}\\
\hline
Layer 6 & 0.328 & 0.390 & 0.421 & 0.433 & 0.435 & 0.435\\
\hline
Layer 6 + landmark retraining & 0.418 & 0.515 & 0.548 & 0.557 & 0.557 & 0.557\\
\hline
Layer 6 + turntable retraining & 0.289 & 0.349 & 0.377 & 0.391 & 0.392 & 0.393\\
\hline
\multicolumn{7}{|c|}{\bf Oxford 105K}\\
\hline
Layer 6 & 0.260 & 0.330 & 0.370 & 0.388 & 0.392 & 0.392\\
\hline
Layer 6 + landmark retraining & 0.354 & 0.467 & 0.508 & 0.523 & 0.524 & 0.522\\
\hline
Layer 6 + turntable retraining & 0.223 & 0.293 & 0.331 & 0.348 & 0.350 & 0.351\\
\hline
\multicolumn{7}{|c|}{\bf Holidays}\\
\hline
Layer 6 & 0.591 & 0.683 & 0.729 & 0.747 & 0.749 & 0.749\\
\hline
Layer 6 + landmark retraining & 0.609 & 0.729 & 0.777 & 0.789 & 0.789 & 0.789\\
\hline
Layer 6 + turntable retraining & 0.587 & 0.702 & 0.741 & 0.756 & 0.756 & 0.756\\
\hline
\multicolumn{7}{|c|}{\bf UKB}\\
\hline
Layer 6 & 2.630 & 3.130 & 3.381 & 3.416 & 3.423 & 3.426\\
\hline
Layer 6 + landmark retraining & 2.410 & 2.980 & 3.256 & 3.297 & 3.298 & 3.300\\
\hline
Layer 6 + turntable retraining & 2.829 & 3.302 & 3.526 & 3.552 & 3.556 & 3.557\\
\hline
\end{tabular}
\vspace{3mm}
\caption{The performance of neural codes (original and retrained) for different PCA-compression rates. The performance of the descriptors is almost unaffected till the dimensionality of $256$ and the degradation associated with more extreme compression is graceful.}
\label{tab:compressed}
\end{table}

The quality of neural codes \Lsi{I} for different PCA compression rates is presented in \tab{compressed}. Overall, PCA works surprisingly well. Thus, the neural codes can be compressed to $256$ or even to $128$ dimensions almost without any quality loss. The advantage of the retrained codes persists through all compression rates. \tab{128compressed} further compares different holistic descriptors compressed to $128$-dimensions, as this dimensionality has been chosen for comparison in several previous works. For Oxford and Holidays datasets, the landmark-retrained neural codes provide a new state-of-the-art among the low-dimensional global descriptors.

\begin{table}[t]
\centering
\addtolength{\tabcolsep}{2pt}
\noindent
\begin{tabular}{|c|c|c|c|c|c|}
\hline
Descriptor & Oxford & Oxford 105K & Holidays & UKB\\
\hline
Fisher+color\cite{Gordo12}           &  ---  &  ---  &  0.723  & 3.08\\
VLAD+adapt+innorm\cite{arandjelovic13}      & 0.448 & 0.374 &  0.625  & --- \\
Sparse-coded features\cite{ge2013sparse} &  ---  &  ---  &  0.727  & \bf{3.67}\\
Triangulation embedding\cite{Jegou14} &  0.433  &  0.353  &  0.617  & 3.40\\
\hline
\multicolumn{5}{|c|}{\bf Neural codes trained on ILSVRC}\\
\hline
Layer 6 & 0.433 & 0.386 & 0.747* & 3.42\\
\hline
\multicolumn{5}{|c|}{\bf After retraining on the Landmarks dataset}\\
\hline
Layer 6 & \bf{0.557} & \bf{0.523} & \bf{0.789*} & 3.29\\
\hline
\multicolumn{5}{|c|}{\bf  After retraining on turntable views (Multi-view RGB-D)}\\
\hline
Layer 6 & 0.391 & 0.348 & 0.756* & 3.55\\
\hline
\end{tabular}
\vspace{3mm}
\caption{The comparison of the PCA-compressed neural codes  (128 dimensions) with the state-of-the-art holistic image descriptors of the same dimensionality. The PCA-compressed landmark-retrained neural codes establish new state-of-the-art on Holidays, Oxford, and Oxford 105K datasets.}
\label{tab:128compressed}
\end{table}

\begin{figure*}
\centering
\begin{tabular}{c}
\includegraphics[width=12cm]{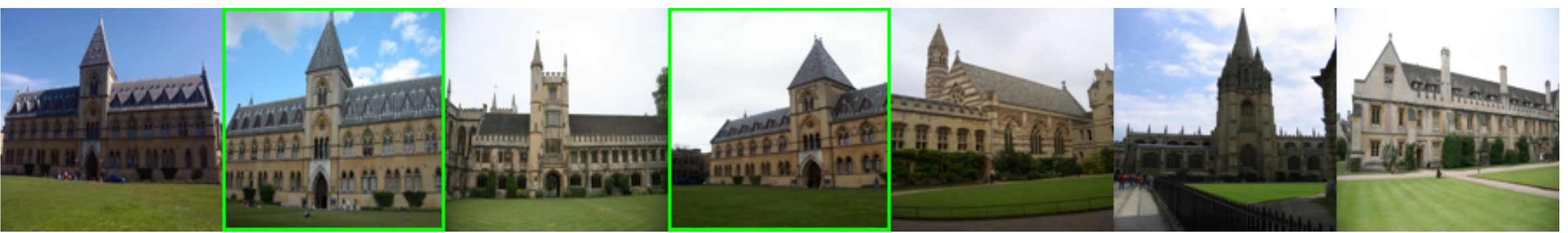}\\
\includegraphics[width=12cm]{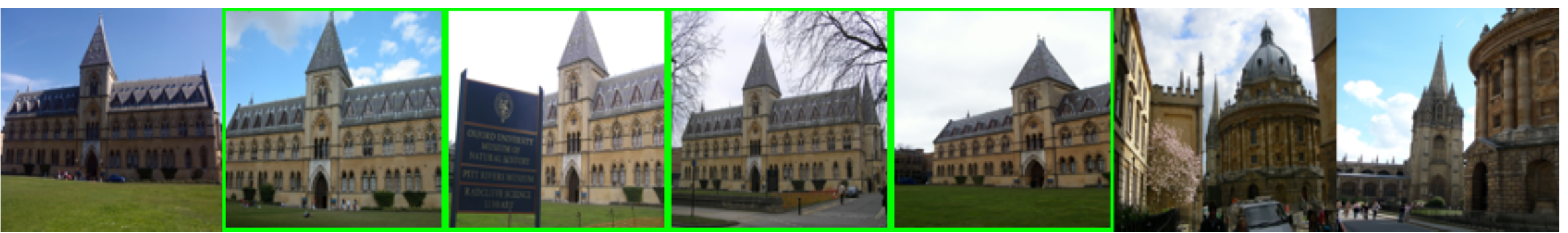}\\
\includegraphics[width=12cm]{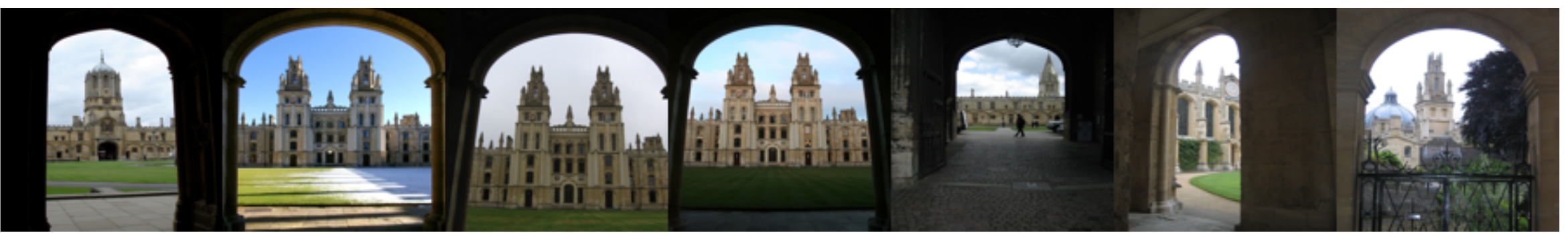}\\
\includegraphics[width=12cm]{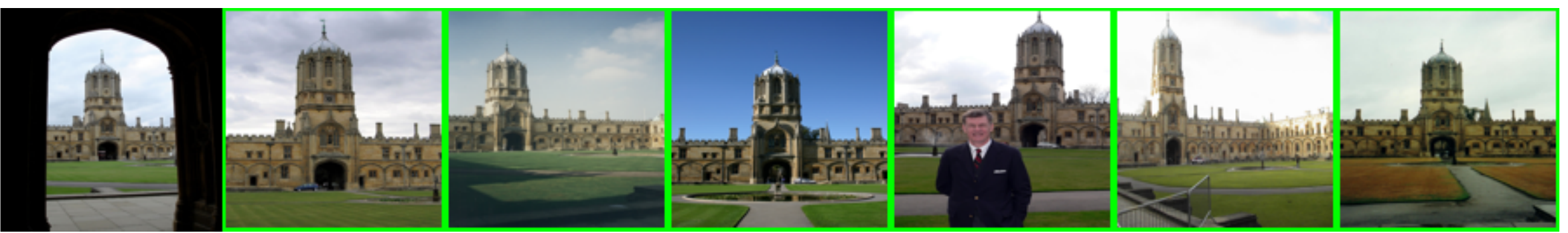}\\
\end{tabular}
\caption{Examples of queries with large differences between the results of the PCA-compressed and the discriminatively-compressed neural codes (for $32$ dimensions). The correct answers are outlined in green.}
\label{fig:adapted_vs_original_32}
\end{figure*}

{\bf Discriminative dimensionality reduction.} In this section, we further perform discriminative dimensionality reduction via the learning of a low-rank projection matrix $W$. The objective of the learning is to make distances between codes small in the cases when the corresponding images contain the same object and large otherwise, thus achieving additional tolerance for nuisance factors, such as viewpoint changes. For such learning, we  collected a number of image pairs which contained the same object. Again, the challenge here was to collect a diverse set of pairs.

\begin{figure*}
\centering
\includegraphics[width=12cm]{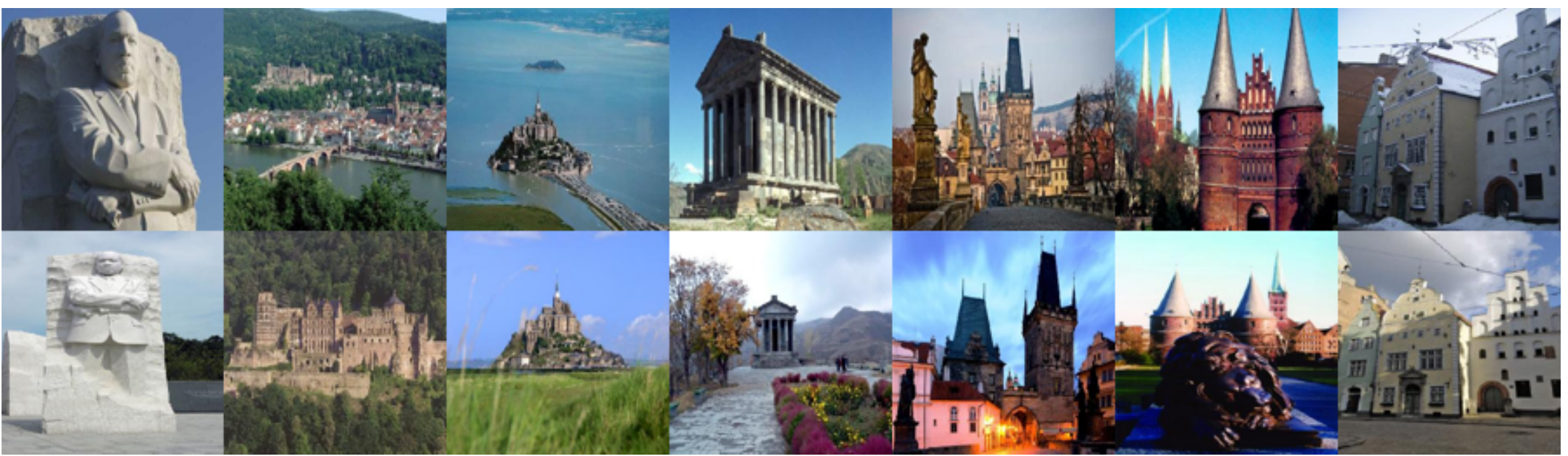}
\caption{Examples of training pairs for discriminative dimensionality reduction. The pairs were obtained through time-consuming RANSAC matching of local features and simple analysis of the resulting match graph (see the text for more details).}
\label{fig:train_pairs}
\end{figure*}

To obtain such a dataset, we sample pairs of images within the same classes of the Landmark dataset. We built a matching graph using a standard image matching pipeline (SIFT+nearest neighbor matching with the second-best neighbor test \cite{Lowe04} + RANSAC validation \cite{Ransac81}). The pipeline is applied to all pairs of images belonging to the same landmark. Once the graph for the landmark is constructed, we took pairs of photographs that share at least one neighbor in the graph but are not neighbors themselves (to ensure that we do not focus the training process on near duplicates). Via such procedure we obtain $900$K diverse image pairs (\fig{train_pairs}). We further greedily select a subset of $100K$ pairs so that each photograph occurs at most once in such a subset, and use this subset of pairs for training.

Given a dataset of matching pairs we learn a linear projection matrix $W$ via the method from \cite{Simonyan13}. In the experiments with large compression rates ($D=16,32$) we project original $4096$-dimensional codes. For the dimensionality $D = 64,\,128$, we observed significantn overfitting due to a large number of parameters within $W$. In this case we first performed PCA-compression to $1024$ dimensions and then learned $W$ for the preliminarily compressed $1024$-dimensional codes.

\begin{table}[t]
\centering
\addtolength{\tabcolsep}{2pt}
\noindent
\begin{tabular}{|c|c|c|c|c|}
\hline
D = & 16 & 32 & 64 & 128\\
\hline
PCA-compression & 0.328 & 0.390 & 0.421 & 0.433\\
Discriminative dimensionality reduction & 0.368 & 0.401 & 0.430 & 0.439\\
\hline
\end{tabular}
\vspace{3mm}
\caption{The comparison of the performances of the PCA compression and a discriminative dimensionality reduction for the original  neural codes on the Oxford dataset. Discriminative dimensionality reduction improves over the PCA reduction, in  particular for the extreme dimensionality reduction.}
\label{tab:metricLearn}
\end{table}

The results of the two compression strategies (PCA and the discriminative reduction) are compared for non-retrained codes for the Oxford dataset in~\tab{metricLearn}. As can be seen, the biggest gain from discriminative dimensionality reduction is achieved for the extremely compressed $16$-dimensional codes. We have also evaluated the discriminative dimensionality reduction on the neural codes retrained on the Landmarks dataset. In this case, however, we did not observed any additional improvement from the discriminative reduction, presumably because the network retraining and the discriminative reduction were performed using overlapping training data.

\section{Discussion}

We have evaluated the performance of the deep neural codes within the image retrieval application. There are several conclusions and observations that one can draw from our experiments.

First of all, as was expected, neural codes perform well, even when one uses the CNN trained for the classification task and when the training dataset and the retrieval dataset are quite different from each other. Unsurprisingly, this performance can be further improved, when the CNN is retrained on photographs that are more related to the retrieval dataset.

We note that there is an obvious room for improvement in terms of the retrieval accuracy, in that all images are downsampled to low resolution ($224 \times 224$) and therefore a lot of information about the texture, which can be quite discriminative, is lost. As an indication of potential improvement, our experiments with Fisher Vectors suggest that their drop in performance under similar circumstances is about $0.03$ mAP on Holidays.

Interestingly, and perhaps unexpectedly, the best performance is observed not on the very top of the network, but rather at the layer that is two levels below the outputs. This effect persists even after the CNN is retrained on related images. We speculate, that this is because the very top layers are too much tuned for the classification task, while the bottom layers do not acquire enough invariance to nuisance factors.

We also investigate the performance of compressed neural codes, where plain PCA or a combination of PCA with discriminative dimensionality reduction result in very short codes with very good (state-of-the-art) performance. An important result is that PCA affects performance of neural codes much less than the one of VLADs, Fisher Vectors, or triangulation embedding. One possible explanation is that passing an image through the network discards much of the information that is irrelevant for classification (and for retrieval). Thus, CNN-based neural codes from deeper layers retain less (useless) information than unsupervised aggregation-based representations. Therefore PCA compression works better for neural codes.

One possible interesting direction for investigation, is whether good neural codes can be obtained directly by training the whole deep architecture using the pairs of matched images (rather than using the classification performance as the training objective), e.g.\ using siamese architecture of \cite{Chopra05}. Automated collection of a suitable training collection having sufficient diversity would be an interesting task on its own. Finally, we note that the dimensionality reduction to a required dimensionality can be realized by choosing the size of a network layer used to produce the codes, rather than a post-hoc procedure.

\clearpage

\bibliographystyle{splncs03}
\bibliography{egbib}
\end{document}